\documentclass[11pt]{article}

\usepackage[margin=1in]{geometry}
\usepackage{fixltx2e,fix-cm}
\usepackage[english]{babel}
\usepackage[T1]{fontenc}
\usepackage{amssymb}
\usepackage{amsmath,amsthm}
\usepackage{graphicx}
\usepackage{makeidx}
\usepackage{multicol}
\usepackage{booktabs}
\usepackage{multirow}
\usepackage{enumitem}
\usepackage{changepage}
\usepackage[caption=false]{subfig}
\usepackage{tikz,xcolor,hyperref}
\usepackage{fancyhdr}

\hypersetup{
    colorlinks=true,
    linkcolor=blue,
    citecolor=blue,
    urlcolor=blue
}

\fancypagestyle{firstpage}{%
  \fancyhf{}
  \fancyfoot[C]{\footnotesize\thepage}
  
}

\definecolor{lime}{HTML}{A6CE39}
\DeclareRobustCommand{\orcidicon}{%
	\begin{tikzpicture}
	\draw[lime, fill=lime] (0,0)
	circle [radius=0.16]
	node[white] {{\fontfamily{qag}\selectfont \tiny ID}};
	\draw[white, fill=white] (-0.0625,0.095)
	circle [radius=0.007];
	\end{tikzpicture}
	\hspace{-2mm}
}

\foreach \x in {A, ..., Z}{%
	\expandafter\xdef\csname orcid\x\endcsname{\noexpand\href{https://orcid.org/\csname orcidauthor\x\endcsname}{\noexpand\orcidicon}}
}

\theoremstyle{plain}

\theoremstyle{definition}

\theoremstyle{remark}

\title{Few-Shot Cross-Dataset Adaptation for Tuberculosis Detection Using DenseNet}

\author{
Bidhan Biswas \\
\and Shahadat Hossain Sohag \\
\and Nabil Ashab \\
\and Soumit Kumar Kundu \\
\and Saif Mahmud Parvez \\
\\
\small Department of Computer Science and Engineering, Dhaka International University, Dhaka, Bangladesh.
}

\date{}

\begin{document}

\thispagestyle{firstpage}

\begin{center}
{\LARGE\bfseries Few-Shot Cross-Dataset Adaptation for Tuberculosis Detection Using DenseNet}\par
\vspace{0.6em}
{\normalsize\itshape Accepted for publication in the Proceedings of the 2026 International Conference on Electrical, Computer and Communication Technologies (ECCT 2026). Dhaka, Bangladesh, May 7–9, 2026.}\par
\vspace{1em}
{\large Bidhan Biswas \quad Shahadat Hossain Sohag \quad Nabil Ashab \quad Soumit Kumar Kundu \quad Saif Mahmud Parvez}\par
\vspace{0.4em}
{\small Department of Computer Science and Engineering, Dhaka International University, Dhaka, Bangladesh.}
\end{center}

\vspace{1em}

\begin{abstract}
Tuberculosis (TB) is one of the most common and dangerous bacterial ailments. Every year, it causes a large number of deaths worldwide. Although many deep learning models can detect tuberculosis from chest X-rays quite accurately, severe domain shift across datasets makes the task challenging. Different imaging protocols, patient demographics, and equipment across domains make the task of generalization difficult. In real-world settings, a model may perform well on one dataset but show a noticeable drop in performance when tested on another. In this work, we address this domain adaptation challenge through a few-shot scaling study. A controlled cross-dataset evaluation is presented in this paper using TBX11K as the source domain and the Mendeley TB dataset as the target domain. It is investigated how varying the number of target samples affects model performance under three training regimes: frozen backbone adaptation, full fine-tuning of a source-pretrained DenseNet121 model, and training from scratch. The results indicate that the model can perform well even with limited data and can achieve 98.36\% accuracy with just 75 labeled samples per class. The adaptation curves demonstrate how fine-tuning effectively mitigates domain shift. These findings establish full fine-tuning of pretrained models as a highly effective and practical strategy for mitigating domain shift in low-resource clinical deployment scenarios.
\end{abstract}

\section{Introduction}
Medical imaging and automation in disease detection from chest X-rays is improving the diagnosis of tuberculosis (TB) more efficient. Early detection of the disease is essential for effective treatment and for limiting its spread. Convolutional neural networks (CNNs) have shown very impressive performance in this area. The real challenge arises when a model trained on one dataset is applied to a different domain. This drop in performance, widely known as domain shift, remains one of the biggest obstacles to actually using these AI tools in everyday clinical practice.

In practice, the early convolutional layers are frozen, which capture basic radiographic patterns such as edges, textures, and general lung structures. These patterns tend to stay consistent across different X-ray machines and hospitals.
It has been shown that updating features and classification layer helps the model to adjust to different contrast, brightness and dataset-specific details.

In this work, we have focused on three major facts:

\begin{itemize}
    \item Initially, it has been observed that a strong pretrained model can lose accuracy significantly on domain shift.
    \item Secondly, the performance can be improved by few-shot adaptation technique on the target dataset. We have carefully checked the performance by increasing the labeled data from 1\% to 15\% of the target dataset and observed the performance improvement.
    \item Finally, the proposed model's performance is evaluated using accuracy, precision, recall, F1-score, and AUC.
\end{itemize}

This work has shown an effective way to work with different clinical settings by applying few-shot learning principles.

\section{Related Work}
In the last few years, deep learning has taken the accuracy level of image-based TB detection to new heights. Different kinds of ensembles, attention mechanisms, and multi-task learning approaches have been proposed to push the performance further. However, most of these models are trained and tested on the same datasets.

Hooda et al.~\cite{hooda2023tuberculosis} ensemble handcrafted texture features with CNN outputs for improved performance. Abideen et al.~\cite{abideen2020uncertainty} experimented with Bayesian Convolutional Neural Networks which can predict uncertainty of the model's predictions. This is a very important feature for clinical applications.

At the same time, some other researchers have worked to classify tuberculosis, COVID-19 and pneumonia from chest X-rays. Ahmed et al.~\cite{ahmed2023joint} developed a model which can classify all three diseases at the same time. Kiche et al.~\cite{kiche2025classification} used data augmentation and oversampling to address class imbalance problems.

Image classification models to work with domain shift have been recognized as a very important challenge. Many models have been developed to address this issue. Musa et al.~\cite{musa2025addressing} advocated for supervised adversarial adaptation to mitigate domain shift. Hwang~\cite{hwang2025domain} showed that MixStyle and multi-level augmentation can generalize the model with higher accuracy.

As it is tough to collect large amounts of labeled data and which is not feasible in many scenarios, many researchers explored few-shot learning approaches. Das et al.~\cite{das2025osltbdnet} proposed such a model using orthogonal softmax layers which is capable of providing a decent performance with a few training samples. Owda et al.~\cite{owda2025lightweight} showed a different hybrid approach which focuses on efficiency even if the computational resources are limited. Devasia et al.~\cite{devasia2023deep} worked on EfficientNet-based~\cite{tan2019efficientnet} models for detecting the affected lung regions.

In this study, we use the TBX11K and Mendeley Tuberculosis Chest X-ray datasets. TBX11K is a large public dataset with high-quality annotations, containing around 11,200 chest X-ray images. In contrast, the Mendeley dataset is smaller, with about 3,000 images, and was collected from a different hospital in Pakistan.

Most previous studies have focused on either zero-shot approaches or fully supervised training on the target dataset. In contrast, this work shows that even without full supervision, a model can achieve promising performance using only a small number of labeled samples from the target dataset.

\section{Datasets}

In this paper, two public tuberculosis chest X-ray datasets have been used: TBX11K and Mendeley TB dataset. Both of these datasets are popular for tuberculosis detection because of the large number of samples and high-quality annotations.
\begin{enumerate}
\item \textbf{TBX11K~\cite{liu2020rethinking}:}
The dataset contains about 11,200 chest X-ray images with proper annotation for normal and tuberculosis cases. The authors of this dataset have collected the data from multiple sources and provided a clear and well-annotated dataset. The dataset contains both classification and localization annotation. In this work, to generalize the model, only the classification labels have been used for training.
The TBX11K dataset consists of multiple categories of TB cases, including healthy, sick but non-TB, active TB and latent TB. All TB-related categories (active and latent) have been treated as positive class while healthy and non-TB cases are grouped as the negative class. This study employs class-aware training strategies and additionally reports evaluation metrics such as F1-score, recall and AUC.
\item \textbf{Mendeley Tuberculosis Chest X-ray~\cite{mendeley2020}:}
This dataset is collected from a local hospital in Pakistan. It contains 2,494 chest X-rays of tuberculosis patients and 514 normal chest X-rays. In this work, the Mendeley dataset has been used as the target dataset. The model is trained on the TBX11K dataset and then tested on the Mendeley dataset to evaluate the performance.

\end{enumerate}
\textbf{Data Handling and Evaluation Protocol:}
The TBX11K dataset is used for initial training which guarantees that each X-ray corresponds to a unique patient. The Mendeley dataset is used for adaptation and evaluation. It has been ensured that all images used for training are strictly separated from those used for testing. In the few-shot setting, a small number of samples per class are selected for training, while all remaining are reserved for evaluation.
The model is trained on TBX11K and evaluated on a completely different dataset (Mendeley) that reduces the risk of unintended data leakage.

In both of the datasets, the X-ray images are classified and verified by expert radiologists. As both of the datasets are collected from different sources, the images have different characteristics in terms of contrast, brightness, and noise levels. This makes the domain shift problem challenging and realistic for real-world clinical deployment.

\section{Methodology}
\subsection{Problem Definition}

The real-world problem that has been addressed in this work is: a well-trained model with decent accuracy on one dataset (TBX11K dataset) can drop its performance significantly when applied to a different dataset (Mendeley TB dataset). Domain shift reduces the model's ability because multiple datasets have different types of imaging characteristics, patient groups and machines.

The goal of this work is to examine how few-shot adaptation techniques can be used to reduce this performance drop by fine-tuning a pretrained model with a small number of labeled samples from the target dataset.

\subsection{Backbone Architecture}

DenseNet-121~\cite{huang2017densely} has been chosen as the backbone architecture for this study. The dense connectivity pattern in DenseNet allows efficient feature reuse with stable gradient flow. This feature of DenseNet is very important for medical imaging where the differences in texture and shadowing can be identified properly. Finally, the fully connected layer of the pretrained network has been replaced with a single node classifier to output the probability of TB presence.

Let $\mathcal{D}_s = \{(x_i^s, y_i^s)\}_{i=1}^{N_s}$ denote the source dataset (TBX11K), where $x_i^s$ represents chest X-ray images and $y_i^s \in \{0,1\}$ corresponds to Normal or TB labels. A deep neural network parameterized by $\theta$ is trained to minimize the empirical risk:

\begin{equation}
\mathcal{L}_s(\theta) = \frac{1}{N_s} \sum_{i=1}^{N_s} \ell \big( f_\theta(x_i^s), y_i^s \big),
\end{equation}

where $\ell(\cdot)$ denotes the cross-entropy loss and $f_\theta(\cdot)$ is the model prediction function. Although this optimization leads to strong performance on $\mathcal{D}_s$, performance degrades when evaluated on a new target dataset $\mathcal{D}_t$ (Mendeley TB dataset), defined as:

\begin{equation}
\mathcal{D}_t = \{(x_j^t, y_j^t)\}_{j=1}^{N_t}.
\end{equation}

The degradation arises because:

\begin{equation}
P_s(x, y) \neq P_t(x, y),
\end{equation}

where $P_s$ and $P_t$ denote the joint data distributions of source and target domains respectively.

To address this limitation, we introduce a few-shot adaptation strategy. Instead of retraining the entire model from scratch on $\mathcal{D}_t$, we fine-tune the pretrained model using a small labeled subset $\mathcal{D}_t^{(k)}$ containing $k\%$ of the target samples:

\begin{equation}
\mathcal{D}_t^{(k)} \subset \mathcal{D}_t, \quad |\mathcal{D}_t^{(k)}| \ll N_t.
\end{equation}

The adaptation objective becomes:

\begin{equation}
\mathcal{L}_t(\theta) = \frac{1}{|\mathcal{D}_t^{(k)}|}
\sum_{(x,y) \in \mathcal{D}_t^{(k)}}
\ell \big( f_\theta(x), y \big).
\end{equation}

\subsection{Training Strategy}

In this paper, the proposed model has been trained with TBX11K~\cite{liu2020rethinking} dataset with Binary Cross-Entropy loss and then Adam optimizer~\cite{kingma2014adam} is used to optimize the performance of the model.
The preprocessing steps are as follows:
\begin{itemize}
    \item Resize all images to 224×224 pixels.
    \item Normalize with ImageNet mean and standard deviation.
    \item Data augmentation using horizontal flip, slight rotation, intensity variation.
\end{itemize}

\subsection{Implementation Details}

All experiments are implemented in PyTorch.

\begin{itemize}
\item Batch size: 16
\item Learning rate: $1 \times 10^{-4}$ (pretraining), $1 \times 10^{-5}$ (fine-tuning)
\item Optimizer: Adam / AdamW
\item Loss function: Binary Cross-Entropy
\item Epochs: 10 (few-shot), 5 (adaptation)
\end{itemize}

\textbf{Reproducibility:}
All experiments are conducted with a fixed random seed (42) to ensure consistent results.

\textbf{Data Augmentation:}
Training data is augmented using horizontal flipping, rotation, and intensity variations
to improve robustness against domain differences.

\subsection{Few-Shot Domain Adaptation Strategy}

Four experimental settings have been implemented to evaluate the effectiveness of few-shot adaptation and they are as follows:

\begin{enumerate}
\item \textbf{Zero-shot Transfer:}
Model trained on TBX11K and directly evaluated on Mendeley without adaptation.

\item \textbf{Frozen Backbone Adaptation:}
Feature extractor is frozen; only the classifier head is updated using $k$ labeled samples per class.

\item \textbf{Full Fine-Tuning:}
All network layers are fine-tuned using $k$ labeled samples per class.

\item \textbf{Training from Scratch:}
Model weights are randomly initialized and trained using only $k$ labeled samples per class.

\end{enumerate}
Few-shot values are defined as:
\[
k \in \{10, 25, 50, 75, 100\}
\]
meaning $k$ labeled images per class are randomly sampled from the target dataset.

\subsection{Algorithm Description}
The overall adaptation procedure is summarized in Algorithm~\ref{alg:fewshot}.

\begin{table}[htbp]
\centering
\caption{Few-Shot Domain Adaptation for TB Detection}
\label{alg:fewshot}
\begin{tabular}{l}
\hline
\textbf{Step 1:} Initialize DenseNet-121 with pretrained weights from TBX11K \\
\textbf{Step 2:} Replace classifier with binary output layer \\
\textbf{Step 3:} Train model on source dataset (TBX11K) \\
\hline
\textbf{For each shot size } $k \in \{10,25,50,75,100\}$: \\
\quad Sample $k$ labeled images per class from target dataset \\
\quad \textbf{If Frozen Backbone Adaptation:} \\
\quad\quad Freeze feature extractor layers \\
\quad\quad Train classifier head \\
\quad \textbf{Else if Full Fine-Tuning:} \\
\quad\quad Unfreeze all layers \\
\quad\quad Fine-tune entire network \\
\quad \textbf{Else if Training from Scratch:} \\
\quad\quad Reinitialize all weights \\
\quad\quad Train using selected samples \\
\quad Evaluate model on target test set \\
\hline
\end{tabular}
\end{table}

\subsection{Experimental Protocol}
The experimental settings are summarized in Table~\ref{tab:settings}.

\begin{table}[ht!]
\centering
\caption{Experimental Settings}
\label{tab:settings}
\begin{tabular}{lccc}
\hline
Setting & Pretrained & Fine-Tuning Scope & Target Labels Used \\
\hline
Zero-shot & Yes & None & 0 \\
Frozen & Yes & Classifier only & $k$ per class \\
Full FT & Yes & All layers & $k$ per class \\
Scratch & No & All layers & $k$ per class \\
\hline
\end{tabular}
\end{table}

\subsection{Challenges Addressed}

Major challenges that have been addressed in this work include:

\begin{enumerate}
    \item Adapting to different clinical settings having different imaging characteristics, patient demographics, and equipment.
    \item Limited availability of labeled target data.
    \item Overcoming the risk of overfitting with limited training data (small $k$)
    \item Class imbalance in clinical datasets.
\end{enumerate}

The above challenges are common in real-world clinical deployment scenarios. The proposed few-shot adaptation strategy is designed to effectively handle these challenges.

\section{Results}

\subsection{Few-Shot Scaling Behavior}

The accuracy of few-shot adaptation model on different settings: keeping the backbone frozen, full fine-tuning, and training from scratch have been shown in Table~\ref{tab:fewshot_results}. Here, the number of labeled target samples have been increased per class from 10 to 100.

\begin{figure}[ht!]
\centering
\includegraphics[width=\linewidth]{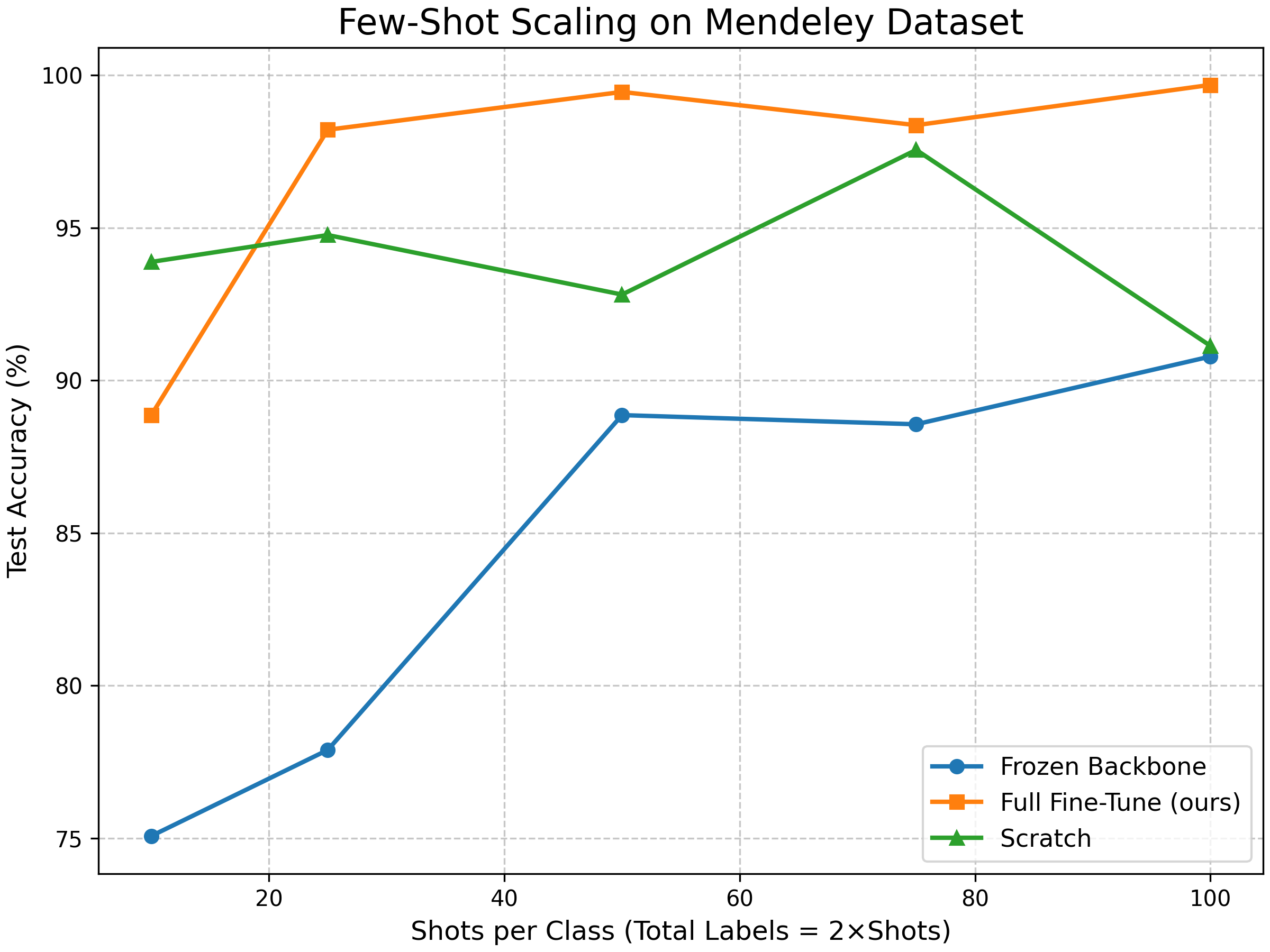}
\caption{Graphically visualizing the performance improvement of few-shot adaptation over frozen backbone and training from scratch.}
\label{fig:scaling}
\end{figure}

\begin{table}[h]
\centering
\caption{Few-shot adaptation accuracy (\%) on the Mendeley dataset.}
\begin{tabular}{c|c|c|c}
\hline
Shots per class & Frozen Backbone & Full Fine-tuning & Scratch \\
\hline
10  & 75.07 & 88.86 & 93.88 \\
25  & 77.89 & 98.21 & 94.76 \\
50  & 88.86 & 99.45 & 92.81 \\
75  & 88.56 & 98.36 & 97.55 \\
100 & 90.78 & 99.68 & 91.13 \\
\hline
\end{tabular}
\label{tab:fewshot_results}
\end{table}

To visualize the scaling behavior clearly, the accuracy numbers have been plotted in Figure~\ref{fig:scaling}.
The following things have been observed from the curves:

\begin{itemize}
    \item Zero-shot does not perform well when applied to different clinical settings.
    \item When more levels are added to frozen-backbone version, it improves the accuracy slowly.
    \item Full fine-tuning shows the fastest improvement and reaches the highest performance level with a few labeled samples.
\end{itemize}

\subsection{Confusion Matrix Analysis}

\begin{figure}[ht!]
\centering
\includegraphics[width=0.8\linewidth]{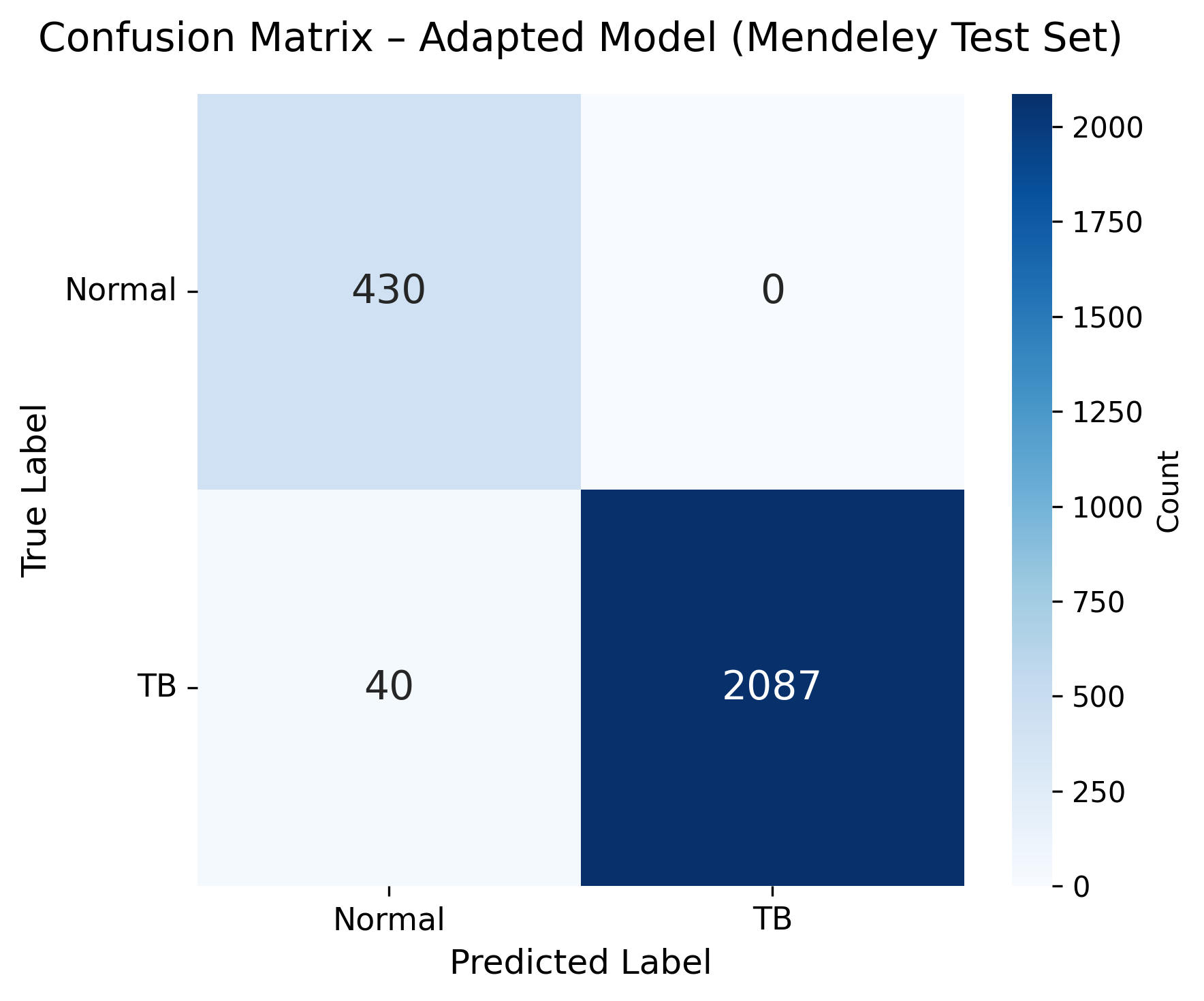}
\caption{Confusion matrix for the full fine-tune model (using the 15\% target adaptation split) on the Mendeley test set. Very few mistakes overall, especially on the TB side.}
\label{fig:cm}
\end{figure}

The output of full fine-tune model has been plotted on the confusion matrix for the Mendeley dataset (Figure~\ref{fig:cm}).

While analyzing the confusion matrix, it has been observed that for the 15\% target adaptation split, the proposed model can correctly identify all 430 normal cases. On the other hand, it successfully identifies 2,087 TB cases out of 2,127 TB cases and failed to detect only about 40 cases. This results in an overall model accuracy of 98.43\%. This is clinically very encouraging because in TB screening, false negatives (missed cases) are usually more concerning than false positives.

\subsection{Describing the model with Grad-CAM}

\begin{figure}[htbp]
\centering

\begin{minipage}{0.48\linewidth}
    \centering
    \includegraphics[width=\linewidth]{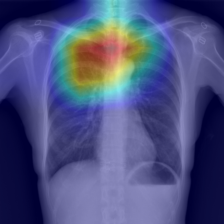}

    (a) Normal case
\end{minipage}
\hfill
\begin{minipage}{0.48\linewidth}
    \centering
    \includegraphics[width=\linewidth]{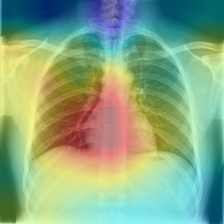}

    (b) Strong attention regions showing TB
\end{minipage}

\caption{Grad-CAM heatmaps for normal and TB cases from the Mendeley dataset}
\label{fig:gradcam}

\end{figure}

The Grad-CAM~\cite{selvaraju2017grad} heatmaps generated on the X-ray images should show which regions the model is paying attention. The Figure~\ref{fig:gradcam} shows two samples from the Mendeley dataset: one normal case and one TB-positive case. In the normal case, the model focuses on the background lung tissue whereas in the TB case, it strongly highlights the central lung abnormalities. These heatmap results show that the model is focusing on clinically relevant areas not only to the image-level characteristics which means that the model is learning correctly.

In the normal example, activation is weak and spread out, mostly on background lung tissue. In the TB example, the model strongly highlights central lung abnormalities which aligns well with radiological expectations for TB. These visualizations increase reliability that the model learns meaningful diagnostic features rather than image artifacts.

\subsection{Scaling Interpretation}

The performance curves with different levels of shots follows a typical saturation pattern which looks like:

\[
\text{Performance} \approx a \left(1 - e^{-b k}\right)
\]

Here,  $k$ is the number of labeled samples per class. The initial spike shows that the pretrained model has absorbed the TB features very well. After the pretraining, only a small amount of target data adjust with the model and it becomes fully fine tuned. The curve again becomes flatten after 50 shots which denotes that the model has reached its performance level near supervised training.

\section{Discussion}

DenseNet can reuse features effectively and this is an important feature for medical image analysis. In this work, the pretrained model on TBX11K with DenseNet-121 builds a strong foundation which has been shown in performance comparison.

The frozen backbone gives a modest improvement with TBX11K feature extraction but when it is applied to the Mendeley dataset, it can not adjust to the domain shift.

On the other hand, full fine-tuning allows the model to adapt with new dataset with only a few labeled samples. It has been shown that with only 75 shots per class, the model can reach 98.36\% accuracy. In the performance comparison graph, it has been observed that the performance curve of full fine-tuning is stepping over the frozen backbone curve.

The confusion matrix also represents the same performance of few-shot fine-tuning model. According to the confusion matrix, the model can correctly identify all 430 normal cases. While detecting TB cases, it misses about 40 TB cases and correctly identifies 2,087 cases. This output is very strong for medical screening because here the false-negatives are more concerning than false positives. The Grad-CAM heatmaps also show that the model is focusing on clinically relevant areas which makes the model reliable also.

The core patterns that have been observed in the paper can be summarized as follows:

\begin{itemize}
    \item In zero-shot setting the model performs poorly in another domain.
    \item Frozen backbone may work modestly with very little data.
    \item Full fine-tuning can achieve excellent performance with limited data.
    \item Scratch training may provide good accuracy but fails when the dataset is large.
\end{itemize}

In a nutshell, it can be said that to deploy the proposed model in real-world clinical settings, firstly the model should be pretrained on a large dataset like TBX11K. Then with a few labeled data from the target domain, the model can be fine-tuned to achieve decent performance.

\begin{table}[ht!]
\centering
\caption{Detailed performance metrics for the adapted model (using the 15\% target adaptation split)}
\label{tab:metrics}
\begin{tabular}{lc}
\toprule
Metric & Value \\
\midrule
Accuracy & 98.43\% \\
Precision (TB) & 100.0\% \\
Recall (TB) & 98.1\% \\
F1-score (TB) & 99.05\% \\
AUC & 99.99\% \\
\bottomrule
\end{tabular}
\end{table}

The model achieves perfect specificity with zero false positives and maintains high recall, demonstrating strong clinical reliability for tuberculosis screening.

\section{Future Work}

This paper opens some new angles of research in the field of domain adaptation for medical image analysis. To detect tuberculosis more effectively, Domain adaptation techniques can be further explored. In Domain adaptation, the model can differentiate between machine artifacts and TB-relevant features. Although this study reports AUC, recall, F1-score, calibration metrics such as calibration error, Brier score are not included. Calibration metrics can be added to make the model more reliable. Additional validation techniques and statistical significance tests can also be incorporated. Another useful direction is to explore parameter-efficient fine-tuning methods, such as adapters or LoRA, which can work well even with limited data and require less computational cost. The proposed model can also be tested on other public datasets to better understand how well it generalizes. New transformer-based architecture can also be explored and their performance can be compared with the DenseNet-121. The model can be extended to multi-class classification to detect other lung diseases like Pneumonia and COVID-19. After analyzing the computational cost and accuracy, the model can be deployed in real-world diagnostic centers and clinics.

\section{Conclusion}

Deep learning models are achieving high accuracy levels in detecting tuberculosis from chest X-rays. In this work, DenseNet-121 has been applied as the backbone architecture and the model has been pretrained on TBX11K dataset. Though the pretrained model performs well on the source dataset, the performance degrades while applying it on the Mendeley dataset.

In real-world perspective it is common to work with different clinical setups. To address this domain adaptation challenge, a few-shot fine-tuning strategy has been applied on the target dataset. The few-shot technique shows that with only 75 labeled samples per class can achieve up to 98.36\% accuracy on the target dataset. The Grad-CAM heatmaps also show that the model is learning clinically relevant features which proves the model's reliability.

\end{document}